\documentclass[a4paper,twocolumn,english,conference]{IEEEtran}
\PassOptionsToPackage{citecolor=gray}{hyperref}
\usepackage{color}
\usepackage{babel}
\usepackage{verbatim}
\usepackage{float}
\usepackage{mathrsfs}
\usepackage{amsmath}
\usepackage{amssymb}
\usepackage{graphicx}
\PassOptionsToPackage{normalem}{ulem}
\usepackage{ulem}
\usepackage[unicode=true,
 bookmarks=true,bookmarksnumbered=true,bookmarksopen=true,bookmarksopenlevel=1,
 breaklinks=false,pdfborder={0 0 0},pdfborderstyle={},backref=false,colorlinks=false]
 {hyperref}
\hypersetup{pdftitle={Your Title},
 pdfauthor={Your Name},
 pdfpagelayout=OneColumn, pdfnewwindow=true, pdfstartview=XYZ, plainpages=false}

\makeatletter

\pdfpageheight\paperheight
\pdfpagewidth\paperwidth

\floatstyle{ruled}
\newfloat{algorithm}{tbp}{loa}
\providecommand{\algorithmname}{Algorithm}
\floatname{algorithm}{\protect\algorithmname}

\usepackage[caption=false,font=footnotesize]{subfig}

\@ifundefined{showcaptionsetup}{}{%
 \PassOptionsToPackage{caption=false}{subfig}}
\usepackage{subfig}
\makeatother

\begin{document}

\title{Clustering For Point Pattern Data}

\author{\IEEEauthorblockN{Quang N. Tran\IEEEauthorrefmark{1}, Ba-Ngu Vo\IEEEauthorrefmark{1},
Dinh Phung\IEEEauthorrefmark{2} and Ba-Tuong Vo\IEEEauthorrefmark{1}}\IEEEauthorblockA{\IEEEauthorrefmark{1}Curtin University, Australia }\IEEEauthorblockA{\IEEEauthorrefmark{2}Deakin University, Australia}}

\maketitle
{\let\thefootnote\relax\footnotetext{Dinh Phung gratefully acknowledges support from the Air Force Office of Scientific Research under award number FA2386-16-1-4138.}}
\begin{abstract}
Clustering is one of the most common unsupervised learning tasks in
machine learning and data mining. Clustering algorithms have been
used in a plethora of applications across several scientific fields.
However, there has been limited research in the clustering of point
patterns \textendash{} sets or multi-sets of unordered elements \textendash{}
that are found in numerous applications and data sources. In this
paper, we propose two approaches for clustering point patterns. The
first is a non-parametric method based on novel distances for sets.
The second is a model-based approach, formulated via random finite
set theory, and solved by the Expectation-Maximization algorithm.
Numerical experiments show that the proposed methods perform well
on both simulated and real data.%
\begin{comment}
{[}{[}NOTE: Abstract: \textasciitilde{}200 words. Paper: max 6 pages,
incl. references.{]}{]}
\end{comment}
\vspace{3mm}
\end{abstract}


\IEEEpeerreviewmaketitle{}

\textbf{\uline{}}

\section{Introduction}

Clustering is a data analysis task that groups similar data items
together \cite{murphy2012machine} and can be viewed as an unsupervised
classification problem since the class (or cluster) labels are not
given \cite{Jain1999data_clustering,russell2003artificial}. Clustering
is a fundamental problem in machine learning with a long history dating
back to the 1930s in psychology \cite{tryon1939cluster}. Today, clustering
is widely used in a host of application areas including genetics \cite{witherspoon2007genetic},
medical imaging \cite{yang2002segmentation}, market research \cite{arimond2001clustering},
social network analysis \cite{mishra2007clustering_social_networks},
and mobile robotics \cite{nguyen2005mobile_robotics}. Excellent surveys
can be found in \cite{Jain1999data_clustering,jain2010clustering50yearsKmeans}. 

Clustering algorithms can broadly be categorized as hard or soft.
In hard clustering, each datum could only belong to one cluster \cite{Jain1999data_clustering,xu2005survey_clustering},
i.e. a hard clustering algorithm outputs a partition of the dataset\footnote{More specifically, the hard clustering could be dichotomized as either
hierarchical or partitional. The hierarchical clustering has the output
as a nested tree of partitions, whereas the output of partitional
clustering is only one partition \cite{murphy2012machine,Jain1999data_clustering}.}. K-means is a typical example of hard clustering. Soft clustering,
on the other hand, allows each datum to belong to more than one cluster
with certain degrees of membership. The Gaussian mixture model \cite{russell2003artificial}
is an example of soft clustering wherein the degree of membership
of a data point to a cluster is given by its mixing probability. Hard
clustering can be obtained from soft clustering results by simply
assigning the cluster with the highest membership degree to each data
point \cite{Jain1999data_clustering}. 

In many applications, each datum is a point pattern, i.e. set or multi-set
of unordered points (or elements). For example, in natural language
processing and information retrieval, the `bag-of-words' representation
treats each document as a collection or set of words \cite{joachims1996probabilistic,mccallum1998comparison_NBtextClassifi}.
In image and scene categorization, the `bag-of-visual-words' representation
\textendash{} the analogue of the `bag-of-words' \textendash{} treats
each image as a set of its key patches \cite{csurka2004visual,fei2005bayesian}.
In data analysis for the retail industry as well as web management
systems, transaction records such as market-basket data \cite{guha1999rock,yang2002clope,yun2001clustering_basket_data}
and web log data \cite{cadez2000EMclustering_VariableLengthData}
are sets of transaction items. Other examples of point pattern data
could be found in drug discovery \cite{dietterich1997solving_multiple_instance},
and protein binding site prediction \cite{minhas2012multiple}. In
multiple instance learning \cite{amores2013multiple_intance_review,foulds2010multi_instance_review},
the `bags' are indeed point patterns. Point patterns are also abundant
in nature, such as the coordinates of trees in a forest, stars in
a galaxy, etc. \cite{Stoyan95,Daley88,Moller03statistical}. 

While point pattern data are abundant, the clustering problem for
point patterns has received very limited attention. To the best of
our knowledge, there are two clustering algorithms for point patterns:
the Bag-level Multi-instance Clustering (BAMIC) algorithm \cite{zhang2009MIClustering};
and the Maximum Margin Multiple Instance Clustering (M$^{3}$IC) algorithm
\cite{zhang2009m3icClustering}. BAMIC adapts the k-medoids algorithm
for the clustering of point pattern data (or multiple instance data)
by using the Hausdorff metric as a measure of dissimilarity between
two point patterns \cite{zhang2009MIClustering}. M$^{3}$IC, on the
other hand, poses the point pattern clustering problem as a non-convex
optimization problem which is then relaxed and solved via a combination
of the Constrained Concave-Convex Procedure and Cutting Plane methods
\cite{zhang2009m3icClustering}. 

In this paper, we propose a non-parametric approach and a model-based
approach to the clustering problem for point pattern data: \vspace{3mm}

\begin{itemize}
\item Our non-parametric approach uses Affinity Propagation (AP) \cite{frey2007_APclustering}
with a novel measure of dissimilarity, known as the Optimal Sub-Pattern
Assignment (OSPA) metric. This metric alleviates the insensitivity
of the Hausdorff metric (used by BAMIC) to cardinality differences.
Moreover, AP is known to find clusters faster and with much lower
error compared to other methods such as k-medoids (used by BAMIC)
\cite{frey2007_APclustering}. \vspace{3mm}
\item In our model-based approach, point patterns are modeled as random
finite sets. Moreover, for a class of models known as independently
and identically distributed (iid) cluster random finite sets, we develop
an Expectation-Maximization (EM) technique \cite{dempster1977em_algorithm}
to learn the model parameters. To the best of our knowledge, this
is the first model-based framework for bag-level clustering of multiple
instance data.\vspace{6mm}
\end{itemize}

\section{Non-parametric clustering for \protect \\
point pattern data\label{sec:AP-clustering}}

In AP, the notion of similarity/dissimilarity between data points
plays an important role. There are various measures of similarity/dissimilarity,
for example distances between observations, joint likelihoods of observations,
or even manually setting the similarity/dissimilarity for each observation
pair \cite{frey2007_APclustering}. In this paper, we use distances
for sets as a measures of dissimilarity. In the following section,
we describe several distances for point patterns. 

\subsection{Set distances\label{subsec:Distance-for-sets}}

Let $X=\left\{ x_{1},...,x_{m}\right\} $ and $Y=\left\{ y_{1},...,y_{n}\right\} $
denote two finite subsets of a metric space $(\mathcal{S},d)$. Note
that when $\mathcal{S}$ is a subset of $\mathbb{R}^{n}$, $d$ is
usually the Euclidean distance, i.e., $d(x,y)=||x-y||$. 

The Hausdorff distance is defined by
\begin{equation}
d_{\mathtt{H}}(X,Y)\triangleq\max\left\{ \max_{x\in X}\min_{y\in Y}d(x,y),\max_{y\in Y}\min_{x\in X}d(x,y)\right\} .\label{eq:Hausdorff_dist}
\end{equation}
and $d_{\mathtt{H}}(X,Y)=\infty$, when $X=\emptyset$ and $Y\neq\emptyset$.

While the Hausdorff distance is a rigorous measure of dissimilarities
between point patterns, it is relatively insensitive to differences
in cardinalities \cite{schuhmacher2008_OSPA,hoffman2004multitarget_distance}.
Consequently, clustering based on this distance has the undesirable
tendency to group together point patterns with large differences in
cardinality.

The Wasserstein distance of order $p\geq1$ is defined by
\begin{equation}
d_{\mathtt{W}}^{(p)}(X,Y)\triangleq\min_{C}\left(\sum_{i=1}^{m}\sum_{j=1}^{n}c_{i,j}d\left(x_{i},y_{j}\right)^{p}\right)^{1/p},\label{eq:OMAT-dist}
\end{equation}
where each $C=\left(c_{i,j}\right)$ is an $m\times n$ transportation
matrix, i.e. $C$ satisfies \cite{schuhmacher2008_OSPA,hoffman2004multitarget_distance}:
\[
\begin{array}{ccc}
c_{i,j}\geq0 & \mbox{ for }1\leq i\leq m, & 1\leq j\leq n\\
\sum_{j=1}^{n}c_{i,j}=\frac{1}{m} & \mbox{ for }1\leq i\leq m,\\
\sum_{i=1}^{m}c_{i,j}=\frac{1}{n} & \mbox{for }1\leq j\leq n.
\end{array}
\]
Note that when $X=\emptyset$ and $Y\neq\emptyset$, $d_{\mathtt{W}}^{(p)}(X,Y)=\infty$
by convention. 

The Wasserstein distance is more sensitive to differences in cardinalities
than the Hausdorff distance and also has a physically intuitive interpretation
when the point patterns have the same cardinality. However, it does
not have a physically consistent interpretation when the point patterns
have different cardinalities, see \cite{schuhmacher2008_OSPA} for
further details. 

The OSPA distance of order $p$ with cut-off $c$ is defined by
\begin{align}
 & d_{\mathtt{O}}^{(p,c)}(X,Y)\triangleq\nonumber \\
 & \left(\frac{1}{n}\left(\min_{\pi\in\Pi_{n}}\sum_{i=1}^{m}\left(\min\left(c,d\left(x_{i},y_{\pi(i)}\right)\right)\right)^{p}+c^{p}\left(n-m\right)\right)\right)^{1/p},\label{eq:OSPA-dist}
\end{align}
if $m\leq n,$ and $d_{\mathtt{O}}^{(p,c)}(X,Y)\triangleq d_{\mathtt{O}}^{(p,c)}(Y,X)$
if $m>n$, where $1\leq p<\infty$, $c>0$, and $\Pi_{n}$ is the
set of permutations on $\left\{ 1,2,...,n\right\} $. The two adjustable
parameters $p$, and $c$, are interpreted as outlier the sensitivity
and cardinality penalty, respectively. The OSPA metric allows for
a physically intuitive interpretation even if the cardinalities of
two sets are not the same, see \cite{schuhmacher2008_OSPA} for further
details.

\subsection{AP clustering with set distances\label{subsec:AP-clustering}}

The AP algorithm first considers all data points as potential exemplars,
i.e., centroids of clusters. Progressively better sets of exemplars
and corresponding clusters are then determined by passing ``responsibility''
and ``availability'' messages based on similarities/dissimilarities
between data points. Further details on the AP algorithm can be found
in \cite{frey2007_APclustering}. The dissimilarities measures considered
in this work are the Hausdorff, Wasserstein, OSPA distances. In the
following section, we will evaluate AP clustering performance with
various set distances.%
\begin{comment}
THESIS or journal version: Put AP pseudo code here. 
\end{comment}

\subsection{Numerical experiments\label{subsec:AP-experiments}}

We evaluate the proposed clustering algorithm with both simulated
and real  data\footnote{For AP clustering algorithm, we use the Frey Lab's code, http://www.psi.toronto.edu/index.php?q=affinity\%20propagation.
Retrieved August 4, 2015.

For computing Hausdorff distance, we use Zachary Danziger's code,
http://www.mathworks.com/matlabcentral/fileexchange/26738. Retrieved
February 1, 2016.

For computing OSPA distance, we use Ba-Ngu Vo's code, http://ba-ngu.vo-au.com/vo/ospa\_dist.zip.
Retrieved February 1, 2016.}. The performance are measured by Rand index \cite{manning2008info_retrieval}.
From a performance perspective, we consider AP clustering with the
Hausdorff metric as version of BAMIC since the only difference is
that latter uses k-medoids instead of AP.

\begin{comment}
comparison with BAMIC and M$^{3}$IC will be done in the journal version
\end{comment}
{} 

In the following experiments, we report the performance of AP clustering
with different set distances, namely Hausdorff, Wasserstein and OSPA.
We use $p=2$ for the Wasserstein and OSPA metrics, and additionally
$c=20$ or $c=60$ for OSPA. 

\subsubsection{AP clustering with simulated data\label{subsec:AP-sim}}

\begin{figure}[h]
\begin{centering}
\vspace{-0mm}
\subfloat[$1^{st}$ dataset]{\begin{centering}
\includegraphics[width=1\columnwidth]{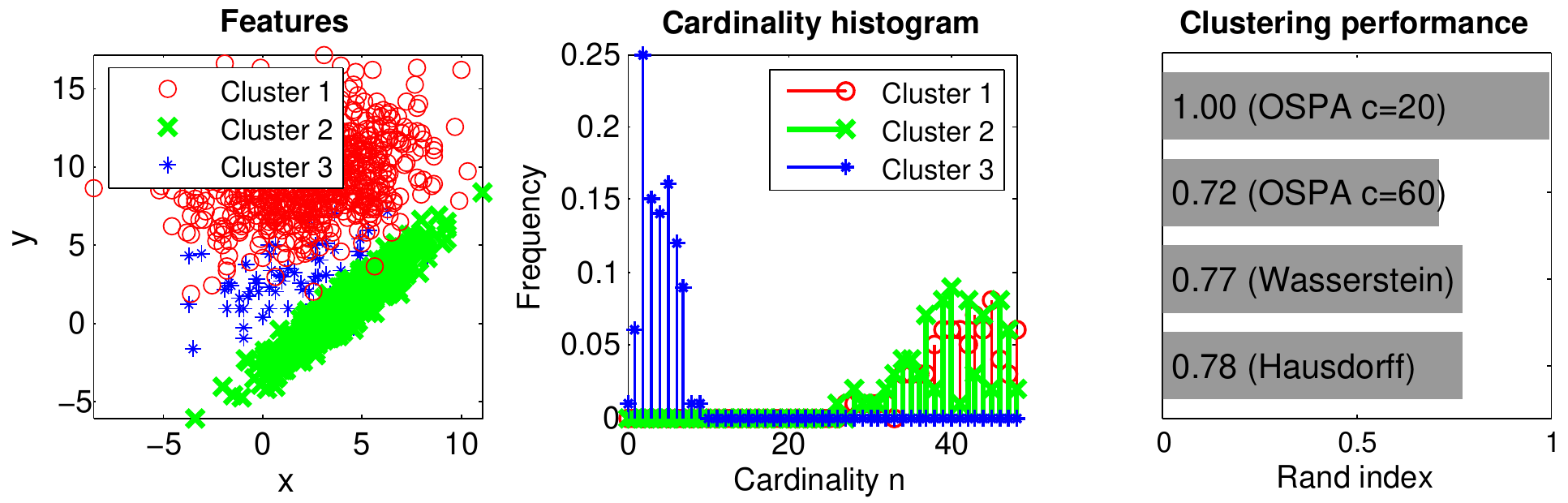}
\par\end{centering}

\vspace{-1mm}
}
\par\end{centering}
\begin{centering}
\subfloat[$2^{nd}$ dataset]{\centering{}\includegraphics[width=1\columnwidth]{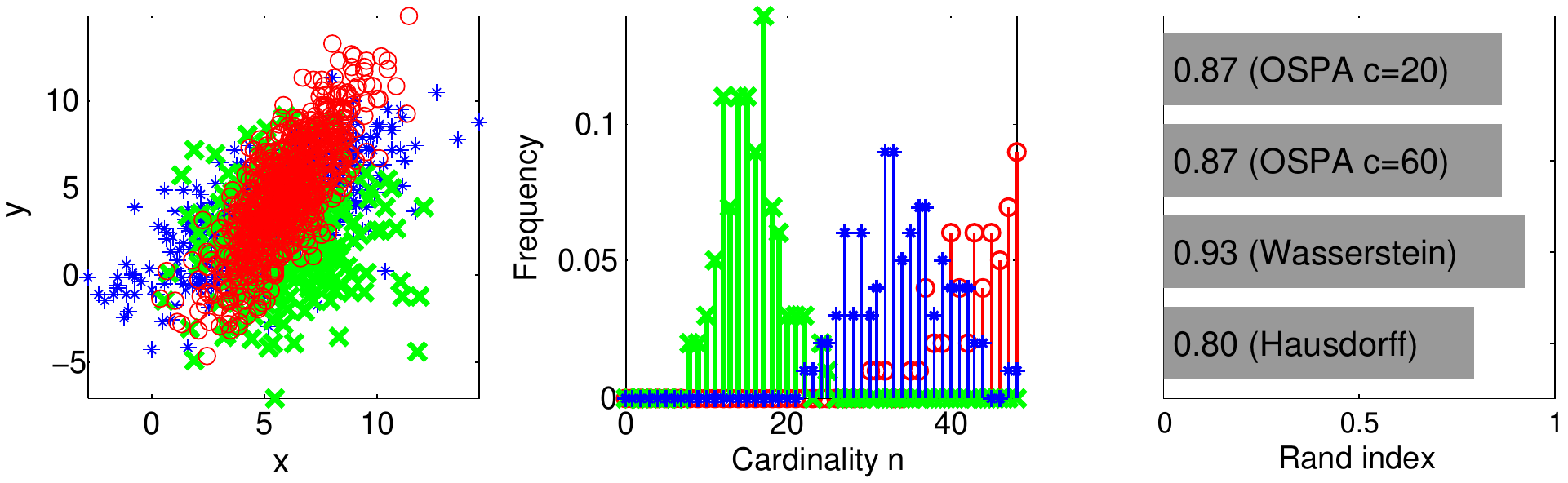}\vspace{-1mm}
}
\par\end{centering}
\begin{centering}
\subfloat[$3^{rd}$ dataset]{\begin{centering}
\includegraphics[width=1\columnwidth]{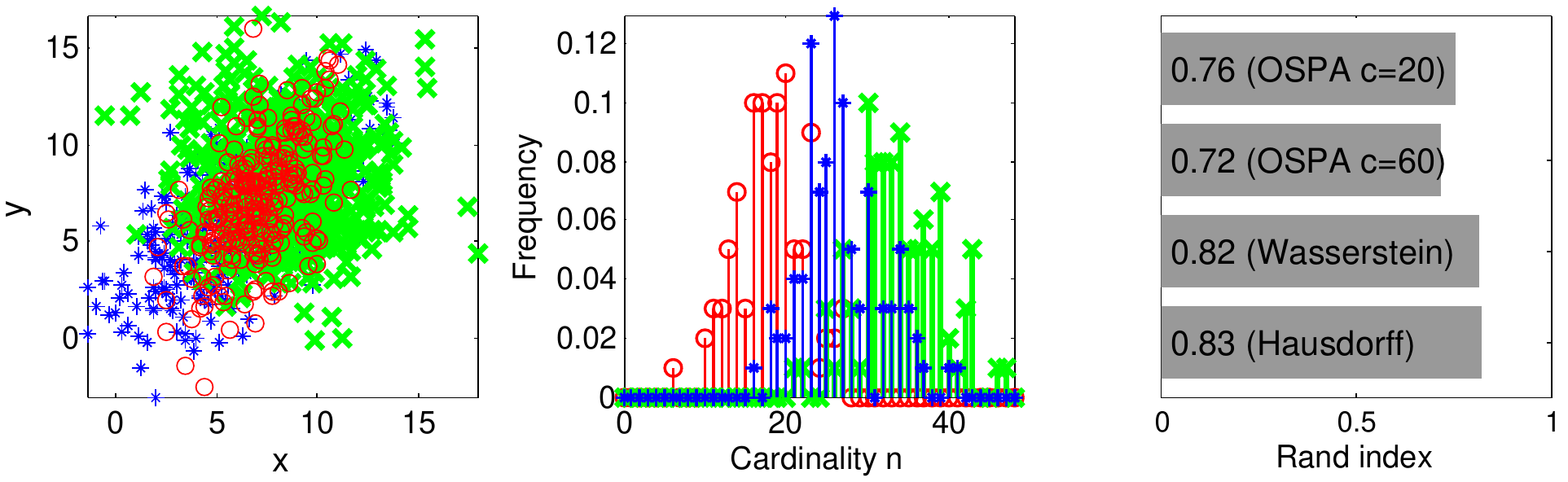}
\par\end{centering}
}
\par\end{centering}
\caption{\label{fig:AP_sim_data}Simulated data from various Poisson RFS distributions.
In each subplot, Left: features of the data, Middle: cardinality histogram
of the data, Right: AP clustering performance with Hausdorff, Wasserstein
and OSPA distances.}

\vspace{-1mm}
\end{figure}

The simulated point pattern data are generated from Poisson RFS models
with 2-D Gaussian feature distributions (see section \ref{subsec:Random-Finite-Set}),
with each Poisson RFS representing a cluster. There are 3 clusters,
each with 100 point patterns, in the simulated data (Fig. \ref{fig:AP_sim_data}). 

\begin{figure}[h]
\begin{centering}
\vspace{-2mm}
\includegraphics[width=0.48\columnwidth]{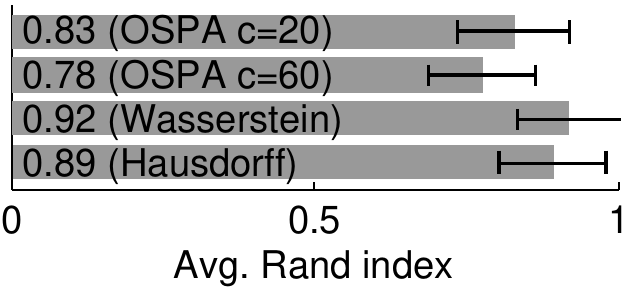}\vspace{-1mm}
\par\end{centering}
\caption{\label{fig:AP_sim_avg_results}Performance of AP clustering with various
set distances on 10 different (simulated) datasets (section \ref{subsec:AP-sim}).
The error-bars are standard deviations of the Rand indices.}
\vspace{-1mm}
\end{figure}

The test is run 10 times with 10 different (simulated) datasets. The
averaged results are shown in Fig. \ref{fig:AP_sim_avg_results},
while individual results for certain datasets are show in Fig. \ref{fig:AP_sim_data}.
Observe that the OSPA's performance could be improved if one choose
a suitable cut-off $c$ (see section \ref{subsec:AP-clustering}).

\subsubsection{AP clustering with real data\label{subsec:AP-Texture}}

\begin{figure}[h]
\begin{centering}
\vspace{-2mm}
\includegraphics[width=0.42\columnwidth]{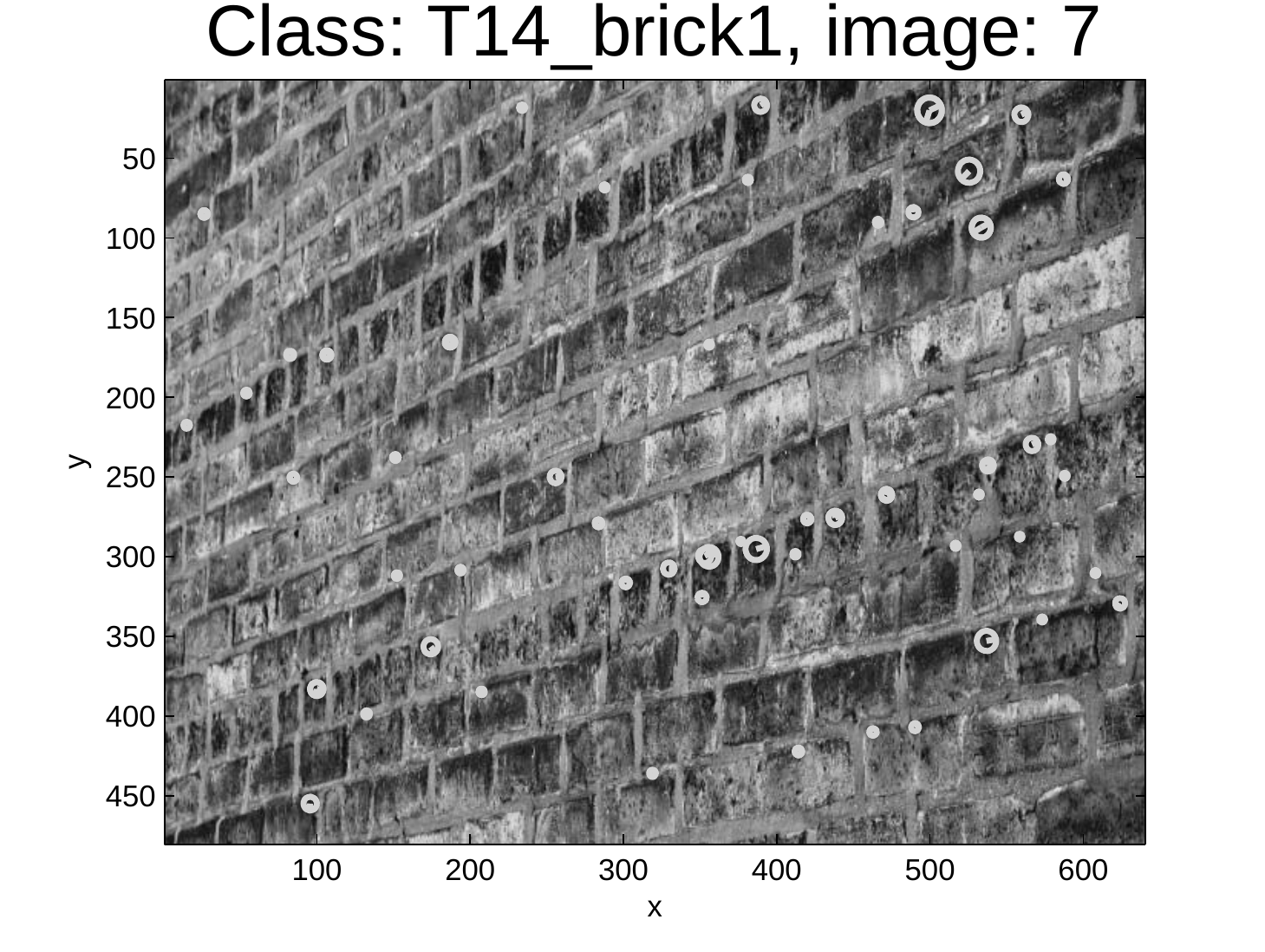}~\includegraphics[width=0.42\columnwidth]{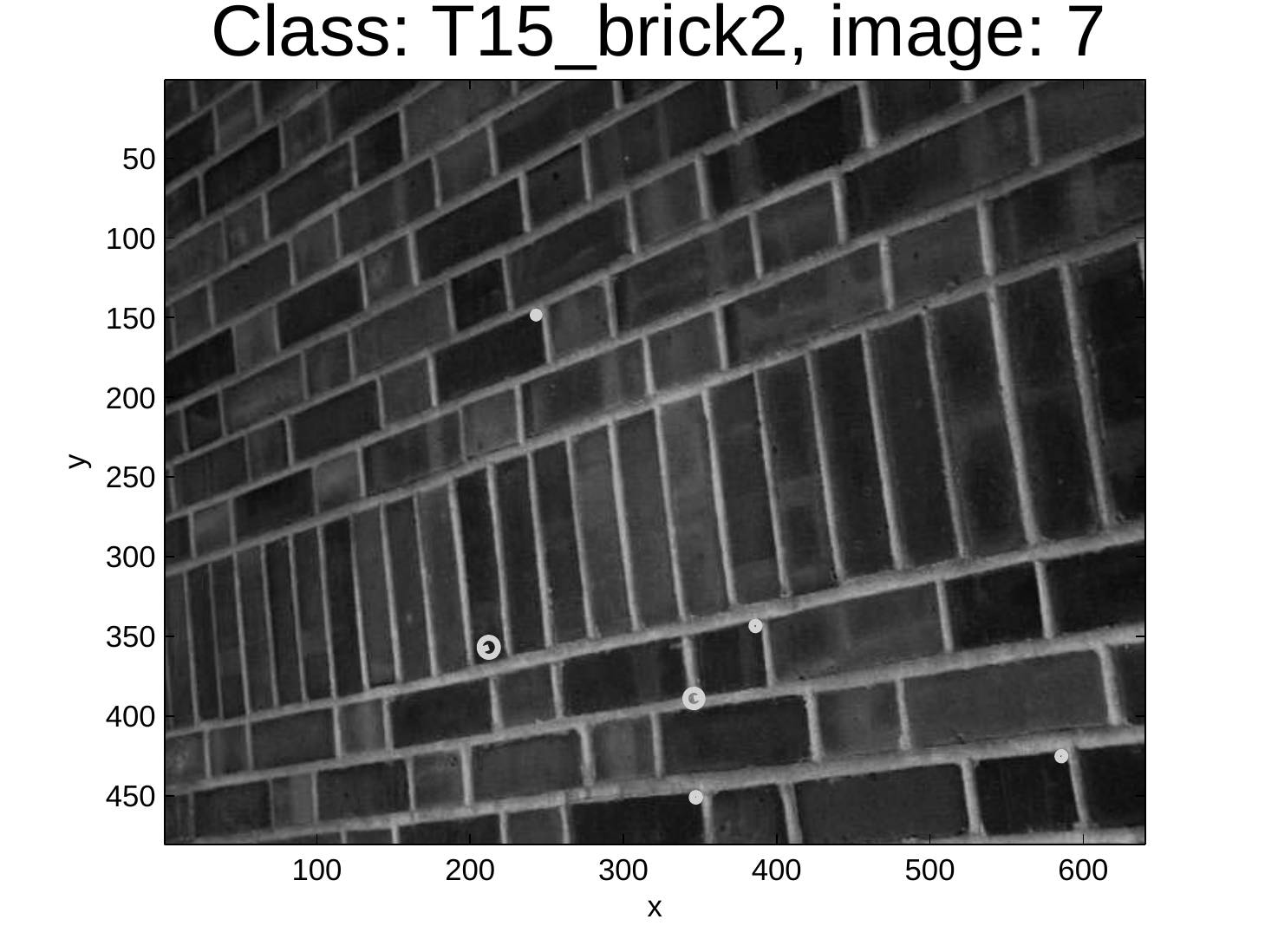}\vspace{-1mm}
\par\end{centering}
\caption{\label{fig:Texture_examples}Example images from classes ``T15\_brick2''
and ``T14\_brick1'' of the Texture dataset. Circles mark detected
SIFT keypoints.}
\vspace{-6mm}
\end{figure}

This experiment involves clustering images from the classes ``T14\_brick1''
and ``T15\_brick2'' of the Texture images dataset \cite{textureDataset}.
Fig. \ref{fig:Texture_examples} visualizes some example images from
these classes. 

\begin{figure}[h]
\begin{centering}
\vspace{-2mm}
\subfloat[]{\centering{}\includegraphics[width=0.42\columnwidth]{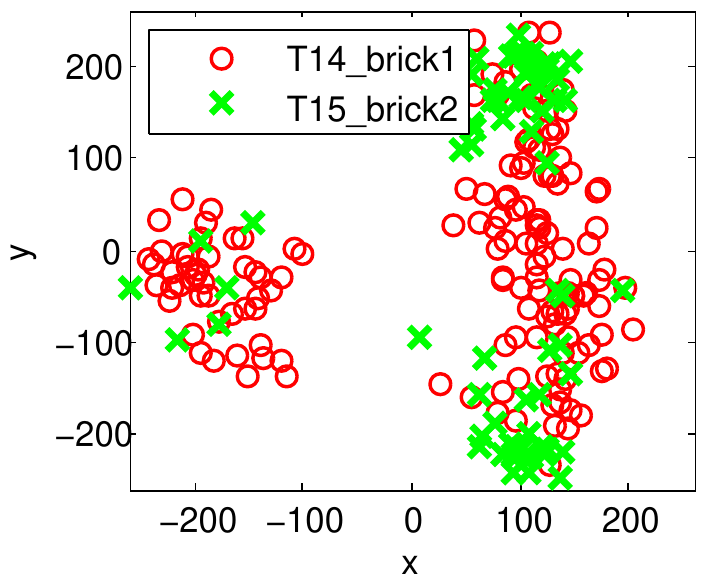}}~\subfloat[]{\begin{centering}
\includegraphics[width=0.42\columnwidth]{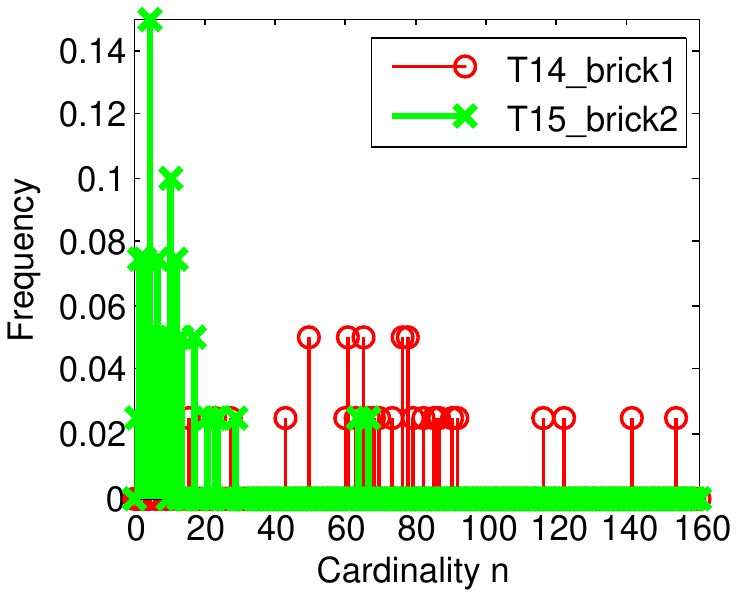}
\par\end{centering}
}
\par\end{centering}
\caption{\label{fig:Texture_data}Extracted data from images of two classes
``T15\_brick2'' and ``T14\_brick1'' of the Texture data. (a) 2-D
features (after applying PCA to the SIFT features) of images. (b)
Histogram of cardinalities of images. }

\vspace{-1mm}
\end{figure}

Features are extracted from each image by applying the SIFT algorithm\footnote{Using the VLFeat library \cite{vlfeatLib}.},
followed by Principal Component Analysis (PCA) to convert the 128-D
SIFT features into 2-D features. Thus each image is compressed into
a point pattern of 2-D features. Fig. \ref{fig:Texture_data} plots
the 2-D features of the images in the Texture images dataset. The
clustering algorithm is used to group together point patterns extracted
from images of the same class.

\begin{figure}[h]
\begin{centering}
\vspace{-1mm}
\includegraphics[width=0.48\columnwidth]{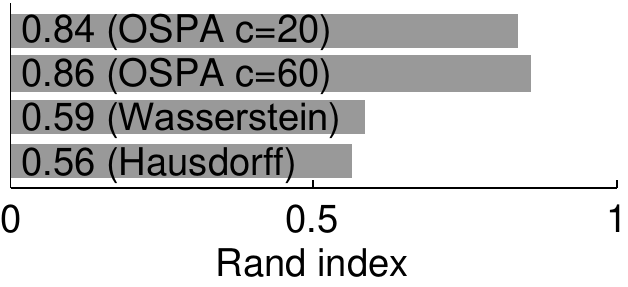}
\par\end{centering}
\vspace{-1mm}

\caption{\label{fig:AP_Texture_results}Performance of AP clustering with various
set distances on the Texture image data set (section \ref{subsec:AP-Texture}).}
\vspace{-4mm}
\end{figure}

The AP clustering results for various set distances are shown in Fig.
\ref{fig:AP_Texture_results}. Observe that for this dataset, the
OSPA distance achieves a better performance than the Hausdorff and
Wasserstein distances.

\section{Model-based clustering for \protect \\
point pattern data\label{sec:EM-clustering}}

In this section, we present generative models for clusters of point
pattern data. In particular, the point patterns are assumed to be
distributed according a finite mixture of iid cluster random finite
set distributions. We then derive an Expectation-Maximization (EM)
algorithm to learn the parameters of these models, which are then
used to cluster the data. 

\subsection{Random Finite Set\label{subsec:Random-Finite-Set}}

Point patterns can be modeled as random finite sets (RFSs), or simple
finite point processes. The likelihood of a point pattern of discrete
features is straightforward since this is simply the product of the
cardinality distribution and the joint probability of the features
given the cardinality. The difficulties arise in continuous feature
spaces. In this work, we only consider continuous feature spaces. 

Let $\mathcal{F}(\mathcal{X})$ denote the space of finite subsets
of a space $\mathcal{X}$. A random finite set (RFS) $X$ of $\mathcal{X}$
is a random variable taking values in $\mathcal{F}(\mathcal{X})$.
An RFS $X$ can be completely specified by a discrete (or categorical)
distribution that characterizes the cardinality $|X|$, and a family
of symmetric joint distributions that characterizes the distribution
of the points (or features) of $X$, conditional on the cardinality.

Analogous to random vectors, the probability density of an RFS (if
it exists) is essential in the modeling of point pattern data. The
probability density $p:\mathcal{F}(\mathcal{X})\rightarrow[0,\infty)$
of an RFS is the Radon-Nikodym derivative of its probability distribution
relative to the dominating measure $\mu$, defined for each (measurable)
$\mathcal{T}$ \ensuremath{\subseteq} $\mathcal{F}(\mathcal{X})$,
by \cite{Stoyan95,Moller03statistical,vo2005sequential,hoang2015cauchy_schwarz}:\vspace{-0mm}
\begin{equation}
\mu(\mathcal{T})=\sum_{m=0}^{\infty}\frac{1}{m!U^{m}}\int\mathbf{1}_{\mathcal{T}}\left(\{x_{1},...,x_{m}\}\right)d\left(x_{1},...,x_{m}\right)\label{eq:commonRefMeasure}
\end{equation}
where $U$ is the unit of hyper-volume in $\mathcal{X}$, and $\mathbf{1}_{\mathcal{T}}(\cdot)$
is the indicator function for $\mathcal{T}$. The measure $\mu$ is
the unnormalized distribution of a Poisson point process with unit
intensity $u=1/U$ when $\mathcal{X}$ is bounded. Note that $\mu$
is unitless and consequently the probability density $p$ is also
unitless. 

In general the probability density of an RFS, with respect to $\mu$,
evaluated at $X=\{x_{1},...,x_{m}\}$ can be written as \cite[p. 27]{lieshout2000markov}
((Eqs. (1.5), (1.6), and (1.7)), \cite{Moller03statistical}:\vspace{-0mm}
\begin{equation}
p(X)=p_{c}(m)\,m!\,U^{m}f_{m}\left(x_{1},...,x_{m}\right),\label{eq:generalRFSdensity}
\end{equation}
where $p_{c}(m)=\mbox{Pr}(|X|=m)$ is the cardinality distribution,
and $f_{m}\left(x_{1},...,x_{m}\right)$ is a symmetric joint probability
density of the points $x_{1},...,x_{m}$ given the cardinality. 

Imposing the independence assumption among the features on the model
in (\ref{eq:generalRFSdensity}) reduces to the \emph{iid-cluster
RFS} model \cite{Daley88}\vspace{-0mm}
\begin{equation}
p(X)=p_{c}(|X|)\,|X|!\,[Up_{f}]^{X}\label{eq:iidRFSdensity}
\end{equation}
where $p_{f}$ is a probability density on $\mathcal{X}$, referred
to as the\emph{ feature density, }and $h^{X}\triangleq\prod_{x\in X}h(x)$,
with $h^{\emptyset}=1$ by convention, is the finite-set exponential
notation.

When $p_{c}$ is a Poisson distribution we have the celebrated \emph{Poisson
point process} (aka, \emph{Poisson RFS})\vspace{-0mm}
\begin{equation}
p(X)=\lambda^{|X|}\,e^{-\lambda}\,[Up_{f}]^{X}
\end{equation}
where $\lambda$ is the mean cardinality. The Poisson model is completely
determined by the intensity function $u=\lambda p_{f}$ \cite{vo2005sequential,hoang2015cauchy_schwarz}.
Note that the Poisson cardinality distribution is described by a single
non-negative number $\lambda$, hence there is only one degree of
freedom in the choice of cardinality distribution for the Poisson
model.

\subsection{Mixture of iid-cluster RFSs\label{subsec:Mixture-of-RFS}}

A mixture of iid-cluster RFSs is a probability density of the form\vspace{-4mm}
\begin{equation}
p(X\mid\Theta)=\sum_{k=1}^{N_{\mathrm{comp}}}w_{k}\,p(X\mid\mathscr{C}_{k},\mathscr{F}_{k})\label{eq:mixture-of-RFSs}
\end{equation}
where $k\in\left\{ 1,...N_{\mathrm{comp}}\right\} $ is the component
label, $w_{k}=\mbox{Pr}(C=k)$ is the component weight (the probability
of an observation belonging to the $k^{th}$ component), and\vspace{-1mm}
\begin{equation}
p(X\mid\mathscr{C}_{k},\mathscr{F}_{k})=p_{c}(|X|\mid\mathscr{C}_{k})\,|X|!\,U^{|X|}\prod_{x\in X}p_{f}(x\mid\mathscr{F}_{k})\label{eq:component-density}
\end{equation}
\vspace{-1mm}
is iid-cluster density of the $k^{th}$ component with cardinality
distribution parameter $\mathscr{C}_{k}$, and feature distribution
parameter $\mathscr{F}_{k}$. 

Note that $\Theta=\left\{ \left(w_{k},\mathscr{C}_{k},\mathscr{F}_{k}\right):k=1,...,N_{\mathrm{comp}}\right\} $
is the complete collection of parameters of the iid-cluster mixture
model. The probability density (\ref{eq:mixture-of-RFSs}) is a likelihood
function for iid point patterns arising from $N_{\mathrm{comp}}$
clusters.

\subsection{EM clustering using mixture of iid-cluster RFSs\label{subsec:EM-clustering}}

\textcolor{black}{Given a dataset $\mathcal{D}=\left(X_{1},...,X_{N_{\mathrm{data}}}\right)$
where each datum is a point pattern $X_{n}=\left\{ x_{n,1},...,x_{n,|X_{n}|}\right\} $.
Assume that $\mathcal{D}$ is generated from an underlying iid-cluster
RFS mixture model with $N_{\mathrm{comp}}$ components (\ref{eq:mixture-of-RFSs})
where each component represents a data cluster. The EM clustering
for the given data includes two main steps:}
\begin{enumerate}
\item \textcolor{black}{\emph{Model learning:}}\textcolor{black}{{} Estimate
the parameters of the underlying model using EM method.}
\item \textcolor{black}{\emph{Cluster assignment:}}\textcolor{black}{{} Assign
observations to clusters using maximum a posteriori (MAP) estimation.}
\end{enumerate}
\textcolor{black}{}
\begin{algorithm}[h]
\textcolor{black}{\vspace{1mm}
}\textbf{\textcolor{black}{\hspace{-2mm}Input}}\textcolor{black}{:
dataset $\mathcal{D}=\left\{ X_{1},...,X_{N_{\mathrm{data}}}\right\} $,
no. of components $N_{\mathrm{comp}}$, no. of iterations $N_{\mathrm{iter}}$
\vspace{1mm}
}

\textbf{\textcolor{black}{Output}}\textcolor{black}{: parameters $\Theta$
of the iid-cluster RFS mixture model\vspace{1mm}
}

\textcolor{black}{initialize $\Theta^{(0)}=\left\{ \left(w_{k}^{(0)},\mathscr{C}_{k}^{(0)},\mathscr{F}_{k}^{(0)}\right):k=1,...,N_{\mathrm{comp}}\right\} $\vspace{1mm}
}

\textbf{\textcolor{black}{for}}\textcolor{black}{{} $i=1$ to $N_{\mathrm{iter}}$\vspace{3mm}
}

\textcolor{black}{~~~}\textsc{\textcolor{black}{\small{}/{*} Compute
posteriors {*}/}}{\small \par}

\textcolor{black}{~~~}\textbf{\textcolor{black}{for}}\textcolor{black}{{}
$n=1$ }\textbf{\textcolor{black}{to}}\textcolor{black}{{} $N_{\mathrm{data}}$ }

\textcolor{black}{~~~~~~}\textbf{\textcolor{black}{for}}\textcolor{black}{{}
$k=1$ }\textbf{\textcolor{black}{to}}\textcolor{black}{{} $N_{\mathrm{comp}}$
\vspace{-1mm}
\begin{equation}
p(k\mid X_{n},\Theta^{(i\mbox{-}1)})=\frac{w_{k}^{(i\mbox{-}1)}\,p(X_{n}\mid\mathscr{C}_{k}^{(i\mbox{-}1)},\mathscr{F}_{k}^{(i\mbox{-}1)})}{\sum_{\ell=1}^{N_{\mathrm{comp}}}{\displaystyle w_{\ell}^{(i\mbox{-}1)}\,p(X_{n}\mid\mathscr{C}_{\ell}^{(i\mbox{-}1)},\mathscr{F}_{\ell}^{(i\mbox{-}1)})}}\label{eq:prior-given-data}
\end{equation}
}\textbf{\textcolor{black}{~~~~~~end }}

\textbf{\textcolor{black}{~~~end }}\textcolor{black}{\vspace{4mm}
}

\textcolor{black}{~~~}\textbf{\textcolor{black}{for}}\textcolor{black}{{}
$k=1$ }\textbf{\textcolor{black}{to}}\textcolor{black}{{} $N_{\mathrm{comp}}$\vspace{1mm}
}

\textcolor{black}{~~~~~~}\textsc{\textcolor{black}{\small{}/{*}
Update component weights {*}/}}\textcolor{black}{\vspace{-2mm}
\begin{equation}
w_{k}^{(i)}=\frac{1}{N_{\mathrm{data}}}\sum_{n=1}^{N_{\mathrm{data}}}p(k\mid X_{n},\Theta^{(i\mbox{-}1)})\,\,\,\,\,\,\,\,\,\,\,\,\,\,\,\,\,\,\,\,\,\,\,\,\,\,\,\,
\end{equation}
}

\textcolor{black}{\vspace{1mm}
~~~~~~}\textsc{\textcolor{black}{\small{}/{*} Update cardinality
distribution parameters {*}/}}{\small \par}

\textcolor{black}{~~~~~~}\textbf{\textcolor{black}{for}}\textcolor{black}{{}
$m=0$ to $N_{\mathrm{card}}$
\begin{equation}
\,\,\,\,\,\,\,\,\,q_{k,m}^{(i)}=\frac{\sum_{n=1}^{N_{\mathrm{data}}}\delta_{(m\mbox{-}|X_{n}|)}\,p(k\mid X_{n},\Theta^{(i\mbox{-}1)})}{\sum_{\ell=0}^{N_{\mathrm{card}}}\sum_{n=1}^{N_{\mathrm{data}}}\delta_{(\ell\mbox{-}|X_{n}|)}\,p(k\mid X_{n},\Theta^{(i\mbox{-}1)})}\label{eq:EM-card}
\end{equation}
~~~~~~}\textbf{\textcolor{black}{end }}

\textcolor{black}{\vspace{2mm}
~~~~~~}\textsc{\textcolor{black}{\small{}/{*} Update feature
density parameters {*}/}}\textcolor{black}{
\begin{align}
\,\,\,\,\,\,\,\,\,\boldsymbol{\mu}_{k}^{(i)}= & \frac{\sum_{n=1}^{N_{\mathrm{data}}}\left(p(k\mid X_{n},\Theta^{(i\mbox{-}1)})\,\sum_{x\in X_{n}}x\right)}{\sum_{n=1}^{N_{\mathrm{data}}}{\displaystyle |X_{n}|}\,p(k\mid X_{n},\Theta^{(i\mbox{-}1)})}\label{eq:EM-feat-mu}\\
\,\,\,\,\,\,\,\,\,\boldsymbol{\Sigma}_{k}^{(i)}= & \frac{\sum_{n=1}^{N_{\mathrm{data}}}{\displaystyle p(k\mid X_{n},\Theta^{(i\mbox{-}1)})}\sum_{x\in X_{n}}\sum_{\ell=1}^{|X_{n}|}\mathbf{K}^{(i)}(x)}{\sum_{n=1}^{N_{\mathrm{data}}}{\displaystyle |X_{n}|\,p(k\mid X_{n},\Theta^{(i\mbox{-}1)})}}\label{eq:EM-feat-cov}
\end{align}
\vspace{-4mm}
}

\textcolor{black}{\hspace*{6mm}where $\mathbf{K}^{(i)}(x)=\left(x-\boldsymbol{\mu}_{k}^{(i)}\right)\left(x-\boldsymbol{\mu}_{k}^{(i)}\right)^{T}$
\vspace{1mm}
}

\textcolor{black}{~~~}\textbf{\textcolor{black}{end }}

\textbf{\textcolor{black}{end }}

\textcolor{black}{return $\Theta^{(N_{\mathrm{iter}})}$}

\textcolor{black}{\caption{\label{alg:EM-for-mix-RFS}EM algorithm for mixture of iid-cluster
RFSs}
}
\end{algorithm}

\textbf{\textcolor{black}{\emph{Model learning step: }}}\textcolor{black}{In
this paper, we present EM learning for the iid-cluster mixture model
(\ref{eq:mixture-of-RFSs}) with categorical cardinality distribution
and Gaussian feature distribution, i.e., the parameters of the $k^{th}$
component are\vspace{-1mm}
\begin{align*}
\mathscr{C}_{k} & =\left\{ \left(q_{k,0},...,q_{k,N_{\mathrm{card}}}\right):0\leq q_{k,m}\leq1,\sum_{m=0}^{N_{\mathrm{card}}}q_{k,m}=1\right\} \\
\mathscr{F}_{k} & =\left\{ \left(\boldsymbol{\mu}_{k},\boldsymbol{\Sigma}_{k}\right)\right\} 
\end{align*}
where $N_{\mathrm{card}}$ is the maximum cardinality of point patterns,
$\boldsymbol{\mu}_{k},\boldsymbol{\Sigma}_{k}$ are the means and
covariances of the Gaussian distributions. The EM algorithm for learning
the model parameters $\Theta=\left\{ \left(w_{k},\mathscr{C}_{k},\mathscr{F}_{k}\right):k=1,...,N_{\mathrm{comp}}\right\} $
proceeds as shown in Algorithm \ref{alg:EM-for-mix-RFS}. }

\textcolor{black}{In each iteration of Algorithm \ref{alg:EM-for-mix-RFS},
the parameters are estimated by $\Theta^{(i)}=\mbox{argmax}_{\Theta}Q\left(\Theta,\Theta^{(i-1)}\right)$
\cite{bilmes1998gentle_EM}, where\vspace{-2mm}
\begin{align}
 & \hspace{-3mm}Q\left(\Theta,\Theta^{(i-1)}\right)\nonumber \\
 & \hspace{-3mm}=\sum_{k=1}^{N_{\mathrm{comp}}}{\displaystyle \sum_{n=1}^{N_{\mathrm{data}}}\log}\left(w_{k}\,p(X_{n}\mid\mathscr{C}_{k},\mathscr{F}_{k})\right)\,p(k\mid X_{n},\Theta^{(i\mbox{-}1)})\label{eq:EM-likelihood}
\end{align}
and $p(X_{n}\mid\mathscr{C}_{k},\mathscr{F}_{k})$, $p(k\mid X_{n},\Theta^{(i\mbox{-}1)})$
are given by (\ref{eq:component-density}), (\ref{eq:prior-given-data})
respectively.}%
\begin{comment}
\textcolor{black}{FOR THESIS (or journal version): provide the proof
of the expressions of parameters.}
\end{comment}
\textcolor{black}{}%
\begin{comment}
\textcolor{black}{FOR THESIS (OR JOURNAL VERSION): }

\textcolor{black}{1. For setting the number of cluster see \cite{dasgupta2000two_round_EM}\cite{xu2005survey_clustering}}

\textcolor{black}{2. Show the log version of EM (log-sum-exp trick).}

\textcolor{black}{3. Discuss use Cauchy Swart divergence to stop EM.}
\end{comment}

\textbf{\textcolor{black}{\emph{Cluster assignment step:}}}\textcolor{black}{\emph{
}}\textcolor{black}{The cluster label $k_{n}$ of an observation $X_{n}$
can be estimated using MAP estimation: \vspace{-1mm}
\begin{equation}
\hat{k}_{n}=\underset{k\in\mathbb{K}}{\mbox{argmax}}p(k\mid X_{n},\Theta),
\end{equation}
where $\mathbb{K}=\left\{ 1,...,N_{\mathrm{comp}}\right\} $, parameters
$\Theta$ are learned by Algorithm \ref{alg:EM-for-mix-RFS}, and
$p(k\mid X_{n},\Theta)$ is the posterior probability:\vspace{-1mm}
\begin{equation}
p(k\mid X_{n},\Theta)=\frac{w_{k}\,p(X_{n}\mid\mathscr{C}_{k},\mathscr{F}_{k})}{\sum_{\ell=1}^{N_{\mathrm{comp}}}{\displaystyle w_{\ell}\,p(X_{n}\mid\mathscr{C}_{\ell},\mathscr{F}_{\ell})}}.
\end{equation}
}

\textcolor{black}{\vspace{-4mm}
}

\subsection{Numerical experiments}

In this section, we evaluate the proposed EM clustering with both
simulated and real data. For comparison with AP clustering, we use
the same datasets from section  \ref{subsec:AP-experiments}. Note
that our experiments assume a mixture of Poisson RFSs model.

\subsubsection{EM clustering with simulated data}

\begin{figure}[h]
\begin{centering}
\vspace{-4mm}
\includegraphics[width=0.48\columnwidth]{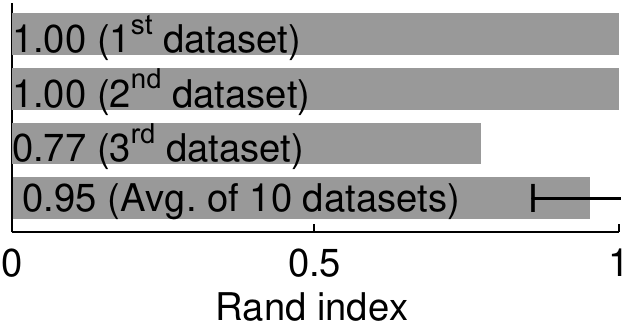}
\par\end{centering}
\caption{\label{fig:EM_sim_performance}Performance of the EM clustering on
simulated datasets (the same datasets from section \ref{subsec:AP-sim}).}
\vspace{-3mm}
\end{figure}

In this experiment we apply EM clustering on the same (simulated)
datasets from section \ref{subsec:AP-sim}. The performance is shown
in Fig. \ref{fig:EM_sim_performance}. EM clustering performs very
well for these datasets with the average Rand index of 0.95.

\begin{figure}[h]
\begin{centering}
\vspace{-3mm}
\subfloat[$1^{th}$ dataset]{\begin{centering}
\includegraphics[width=0.8\columnwidth]{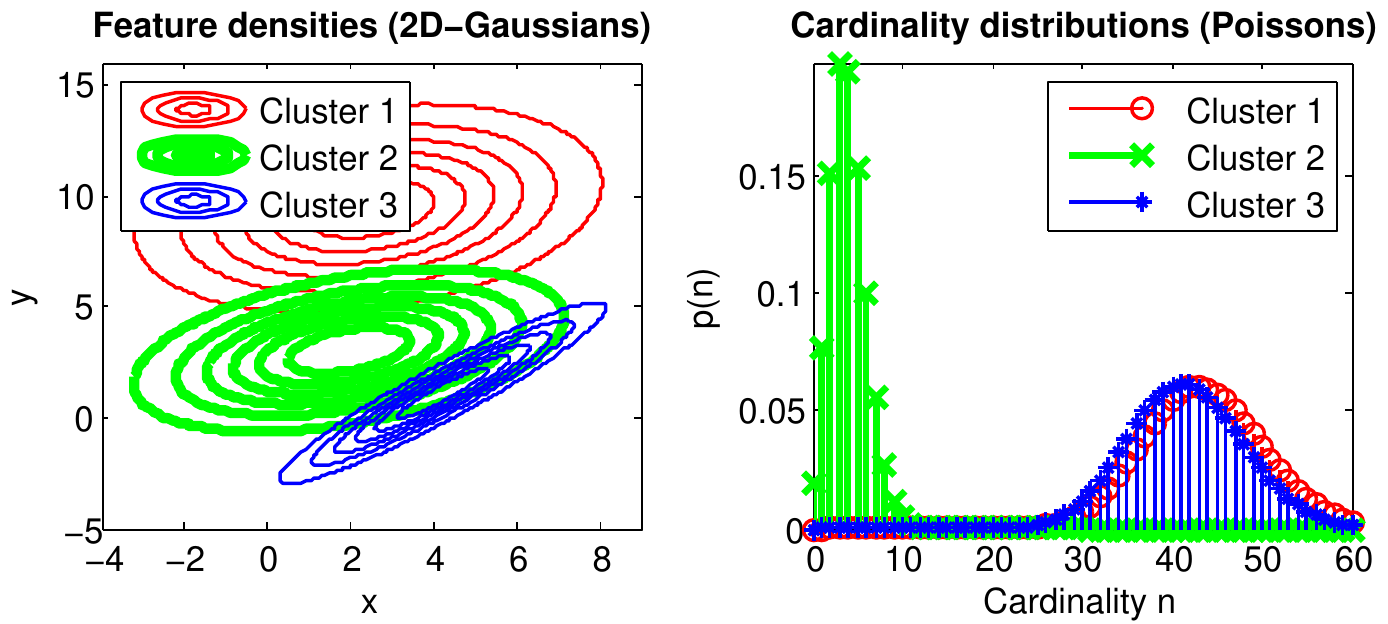}
\par\end{centering}
}
\par\end{centering}
\begin{centering}
\subfloat[$2^{th}$ dataset]{\centering{}\includegraphics[width=0.8\columnwidth]{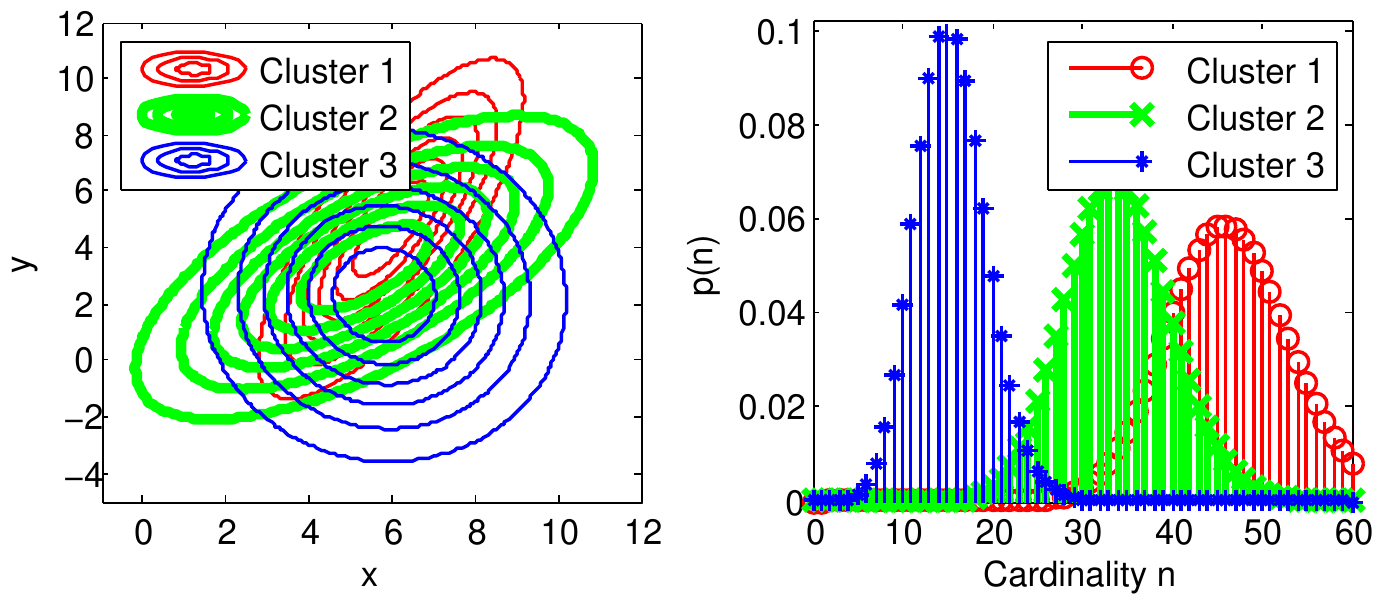}}
\par\end{centering}
\begin{centering}
\subfloat[$3^{th}$ dataset]{\begin{centering}
\includegraphics[width=0.8\columnwidth]{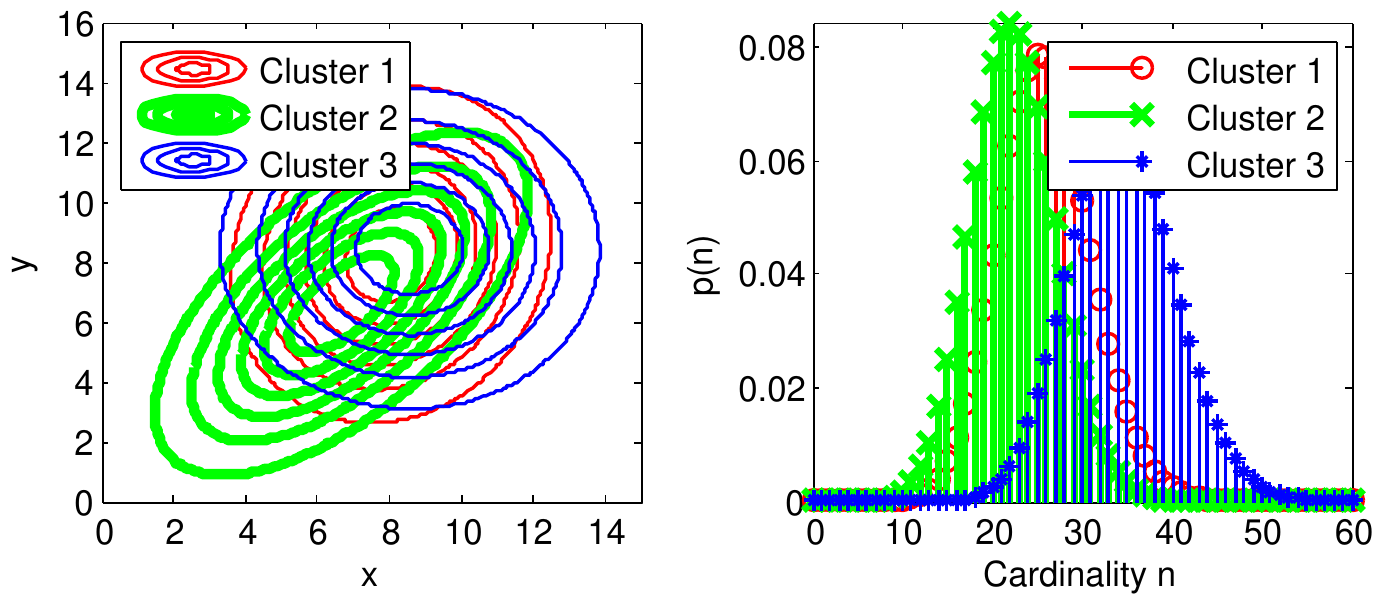}
\par\end{centering}
}
\par\end{centering}
\caption{\label{fig:EM_sim_dist}The RFS distributions learned by EM for 3
simulated datasets (corresponding with 3 datasets in Fig. \ref{fig:AP_sim_data}).
In each subplot, Left: feature distributions, Right: cardinality distributions.}

\vspace{-2mm}
\end{figure}

To further investigate performance of the proposed algorithm, we illustrate
in Fig. \ref{fig:EM_sim_dist} the distributions learned by the EM
algorithm for the 3 datasets shown in Fig. \ref{fig:AP_sim_data}.
Observe that the model-based method has better performance than the
non-parametric method when there is little overlap between the feature
distributions or cardinality distributions of the data model (e.g.,
the $1^{st}$ and $2^{nd}$ simulated dataset). However, as expected,
if there is significant overlap in both the feature distributions
and the cardinality distributions, then model-based clustering performs
poorly since there is not enough information to separate the data
(e.g., the $3^{rd}$ simulated dataset). 

\begin{center}
\begin{figure}[h]
\begin{centering}
\vspace{-0mm}
\includegraphics[width=0.51\columnwidth]{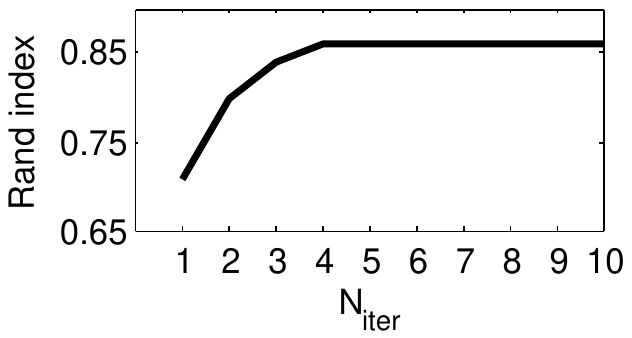}\vspace{-2mm}
\par\end{centering}
\caption{\label{fig:EM_Texture_performance}Performance of the EM clustering
on the Texture dataset (described in section \ref{subsec:AP-Texture}).
The clustering is evaluated with different number of iterations $N_{\mathrm{iter}}=\{1,...,10\}$.}
\end{figure}
\par\end{center}

\subsubsection{EM clustering with real data}

\begin{figure}[h]
\begin{centering}
\vspace{-0mm}
\includegraphics[width=0.8\columnwidth]{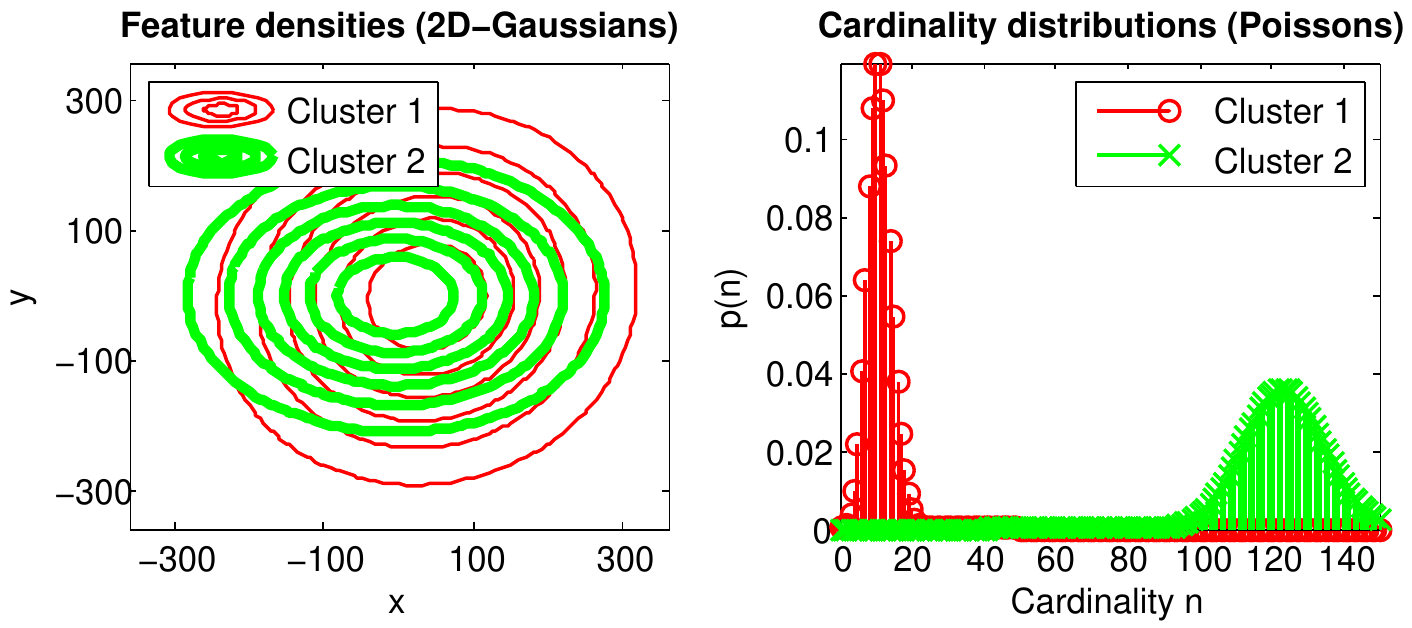}
\par\end{centering}
\caption{\label{fig:EM_Texture_dist}The learned RFS distributions for Texture
data after 10 EM iterations. Left: feature distributions, Right: cardinality
distributions.}
\vspace{-6mm}
\end{figure}

In this experiment, we cluster the Texture dataset (described in section
\ref{subsec:AP-Texture}) using EM clustering with various iterations
$N_{\mathrm{iter}}=\{1,...,10\}$. The performance is shown in Fig.
\ref{fig:EM_Texture_performance}. The best performance is Rand index
of 0.86, which is equal the best performance of AP clustering using
OSPA with $c=60$ (Fig. \ref{fig:AP_Texture_results}). Fig. \ref{fig:EM_Texture_dist}
shows the learned distribution after 10 EM iterations. 

\section{Conclusion}

This paper has detailed a non-parametric approach (based on set distance)
and a model-based approach (based on random finite set) to the clustering
problem for point pattern data (aka `bags' or multiple instance data).
Experiments with both simulated and real data indicate that, in the
non-parametric method, the choice of distance has a big influence
on the clustering performance. The experiments also indicate that
the model-based method has better performance than the non-parametric
method when there is little overlap between the feature distributions
or cardinality distributions of the data model. However, as expected,
if there is significant overlap in both the feature distributions
and the cardinality distributions, then model-based clustering performs
poorly since there is not enough discriminative information to separate
the data. 

\textcolor{black}{Future research directions may include developing
more complex RFS models such as RFSs with Gaussian mixture feature
distribution which can capture better multiple modes feature data
(e.g., section \ref{subsec:AP-Texture}). Another promising development
is adapting the proposed clustering algorithms into data stream mining
\textendash{} an emerging research topic dealing with rapidly and
continuously generated data such as search or surveillance data \cite{nguyen2015stream_clustering}. }

\vspace{-2mm}

\bibliographystyle{IEEEtran}
\bibliography{Refs/ref1}
\vspace{-2mm}

\end{document}